\documentclass{article}
\usepackage{ijcai17}
\usepackage{times}
\usepackage{amsmath,amssymb,amsfonts,bm}
\usepackage[ruled]{algorithm2e}
\usepackage{graphicx}
\usepackage{epstopdf}
\usepackage[tight,normal]{subfigure}
\usepackage{multirow,makecell}
\usepackage{url,footmisc}

\begin{document}

\title{Correlation Hashing Network for Efficient Cross-Modal Retrieval}


\author{
    Yue Cao$^{\dag}$, Mingsheng Long$^{\dag}$, Jianmin Wang$^{\dag}$ \and Philip S. Yu$^{\dag\ddag}$ \\
$^\dag$School of Software, Tsinghua National Laboratory (TNList), Tsinghua University, Beijing, China\\
$^\ddag$Institute for Data Science, Tsinghua University \& University of Illinois at Chicago, IL, USA\\
    {\tt\small caoyue10@gmail.com, \{mingsheng,jimwang\}@tsinghua.edu.cn, psyu@uic.edu}
}

\maketitle

\begin{abstract}
Hashing is widely applied to approximate nearest neighbor search for large-scale multimodal retrieval with storage and computation efficiency. Cross-modal hashing improves the quality of hash coding by exploiting semantic correlations across different modalities. Existing cross-modal hashing methods first transform data into low-dimensional feature vectors, and then generate binary codes by another separate quantization step. However, suboptimal hash codes may be generated since the quantization error is not explicitly minimized and the feature representation is not jointly optimized with the binary codes. This paper presents a Correlation Hashing Network (CHN) approach to cross-modal hashing, which jointly learns good data representation tailored to hash coding and formally controls the quantization error. The proposed CHN is a hybrid deep architecture that constitutes a convolutional neural network for learning good image representations, a multilayer perception for learning good text representations, two hashing layers for generating compact binary codes, and a structured max-margin loss that integrates all things together to enable learning similarity-preserving and high-quality hash codes. Extensive empirical study shows that CHN yields state of the art cross-modal retrieval performance on standard benchmarks.
\end{abstract}

\section{Introduction}\label{section:Introduction}
While large-scale, high-dimensional multimedia big data are pervasive in search engines and social networks, cross-modal retrieval has attracted increasing attention, which enables approximate nearest neighbors (ANN) search across different modalities with computation efficiency and search quality. As relevant data from different modalities (image and text) may endow semantic correlations, it is important to support cross-modal retrieval that returns semantically-relevant results of one modality in response to a query of different modality. A promising solution to the cross-modal retrieval is hashing methods \cite{cite:Arxiv14HashSurvey}, which transform high-dimensional data into compact binary codes and generate similar binary codes for similar data. This paper focuses on cross-modal hashing that builds data-dependent hash coding for efficient cross-media retrieval \cite{cite:TPAMI14Wiki}. Due to large volumes and the semantic gap \cite{cite:TPAMI00SemanticGap}, effective cross-modal hashing remains a challenge.

Existing cross-modal hashing methods construct correlation across different modalities in the process of hash function learning and indexes cross-modal data into an isomorphic Hamming space \cite{cite:CVPR10CMSSH,cite:IJCAI11CVH,cite:NIPS12CRH,cite:KDD12MLBE,cite:SIGMOD13IMH,cite:VLDB14MSAE,cite:SIGIR14DCDH,cite:CVPR14CH,cite:AAAI14SCM,cite:IJCAI15QCH,cite:SIGIR16CCQ}. They can be categorized into unsupervised methods and supervised methods. While unsupervised methods are general and can be trained without semantic labels or relevance feedbacks, they are restricted by the semantic gap \cite{cite:TPAMI00SemanticGap} that high-level semantic description of an object differs from low-level feature descriptors. Supervised methods can incorporate semantic labels or relevance feedbacks to mitigate the semantic gap \cite{cite:TPAMI00SemanticGap} and improve the hashing quality, i.e. achieve accurate search with shorter codes. 

Recently, deep hashing methods \cite{cite:AAAI14CNNH,cite:CVPR15DNNH} have shown that both feature representation and hash coding can be learned more effectively using deep neural networks \cite{cite:NIPS12CNN,cite:ICLR14NIN}, which can naturally encode nonlinear hashing functions. Other cross-modal retrieval models via deep learning \cite{cite:TPAMI14MMNN,cite:JMLR14MMDL,cite:VLDB14MSAE,cite:MM14DL,cite:ARXIV16DCMH} have shown that deep models can capture nonlinear cross-modal correlations more effectively and yielded state-of-the-art results on many benchmarks. However, a crucial disadvantage of these cross-modal deep hashing methods is that the quantization error is not statistically minimized hence the feature representation is not optimally compatible with binary hash coding. Another potential limitation is that they generally do not adopt principled pairwise loss function to link the pairwise Hamming distances with the pairwise similarity labels which is crucial to close the gap between the Hamming distance on binary codes and the metric distance  on continuous representations. Therefore, suboptimal representation and hash coding may be produced by existing cross-modal deep hashing methods.

This paper presents Correlation Hashing Network (CHN), a hybrid deep architecture for cross-modal hashing. CHN jointly learns good image and text representations tailored to hash coding and formally controls the quantization error, which constitutes four components: (1) an image network with multiple convolution-pooling layers to extract good image representations, and a text network with multiple fully-connected layers to extract good text representations; (2) two hashing layers to generate compact hash codes for each modality; (3) a cosine max-margin loss for capturing cross-modal correlation structure; and (4) a new quantization max-margin loss for controlling the quality of the binarized hash codes. Extensive experiments show that CHN yields state-of-the-art results on standard cross-modal retrieval datasets. 


\section{Related Work}\label{section:RelatedWork}
Cross-modal hashing has been a popular research topic in machine learning, computer vision, and multimedia retrieval \cite{cite:CVPR10CMSSH,cite:IJCAI11CVH,cite:NIPS12CRH,cite:KDD12MLBE,cite:SIGMOD13IMH,cite:VLDB14MSAE,cite:SIGIR14DCDH,cite:MM14IMVH,cite:AAAI14SCM,cite:CVPR14CH,cite:SIGIR16CCQ}. We refer readers to \cite{cite:Arxiv14HashSurvey} for a comprehensive survey.

Existing cross-modal hashing methods can be categorized into unsupervised methods and supervised methods. IMH \cite{cite:SIGMOD13IMH} and CVH \cite{cite:IJCAI11CVH} are unsupervised methods that extend spectral hashing \cite{cite:NIPS09SH} to multimodal data. CMSSH \cite{cite:CVPR10CMSSH}, SCM \cite{cite:AAAI14SCM} and QCH \cite{cite:IJCAI15QCH} are supervised methods, which require that if two points are known to be similar, then their corresponding hash codes from different modalities should be made similar. Since supervised methods can exploit semantic labels or relevance information to distill cross-modal correlation and reduce semantic gap \cite{cite:TPAMI00SemanticGap}, they can achieve superior accuracy than unsupervised methods for cross-modal similarity search with shorter hash codes.

Prior cross-modal hashing methods based on shallow architectures cannot effectively exploit the correlation across different modalities. Deep multimodal embedding methods \cite{cite:NIPS13Devise,cite:CVPR15LRCN} have shown that deep models can bridge heterogeneous modalities more effectively for image description. Recent deep hashing methods \cite{cite:AAAI14CNNH,cite:CVPR15DNNH,cite:AAAI16DHN} have given state of the art results, but they can only be used for single-modal retrieval. To our knowledge, Deep Visual-Semantic Hashing (DVSH) \cite{cite:KDD16DVSH} and Deep Cross-Modal Hashing (DCMH) \cite{cite:ARXIV16DCMH} are the only two cross-modal deep hashing methods that use deep networks for representation learning and hash coding. However, our method shares the same problem setting with DCMH that only requires similarity labels across images and texts, while DVSH further requires bimodal image-text pairs to learn \emph{modal-shared} representation. As the most similar work to ours, DCMH adopts inner product between continuous representations as the approximation to the Hamming distance between binary codes, which is not appropriate since the former takes values in $(-\infty,+\infty)$ while the latter takes values in $[-b,+b]$ ($b$ is the number of bits). Furthermore, DCMH adopts Iterative Quantization (ITQ) \cite{cite:CVPR11ITQ} to generate binary codes, which may be not robust to outlier bits when the codes are unbalanced. Our CHN jointly maximizes cross-modal correlation and controls quantization error in a hybrid deep architecture with well-specified loss functions.

\begin{figure*}[tbp]
  \centering
  \includegraphics[width=1.0\textwidth]{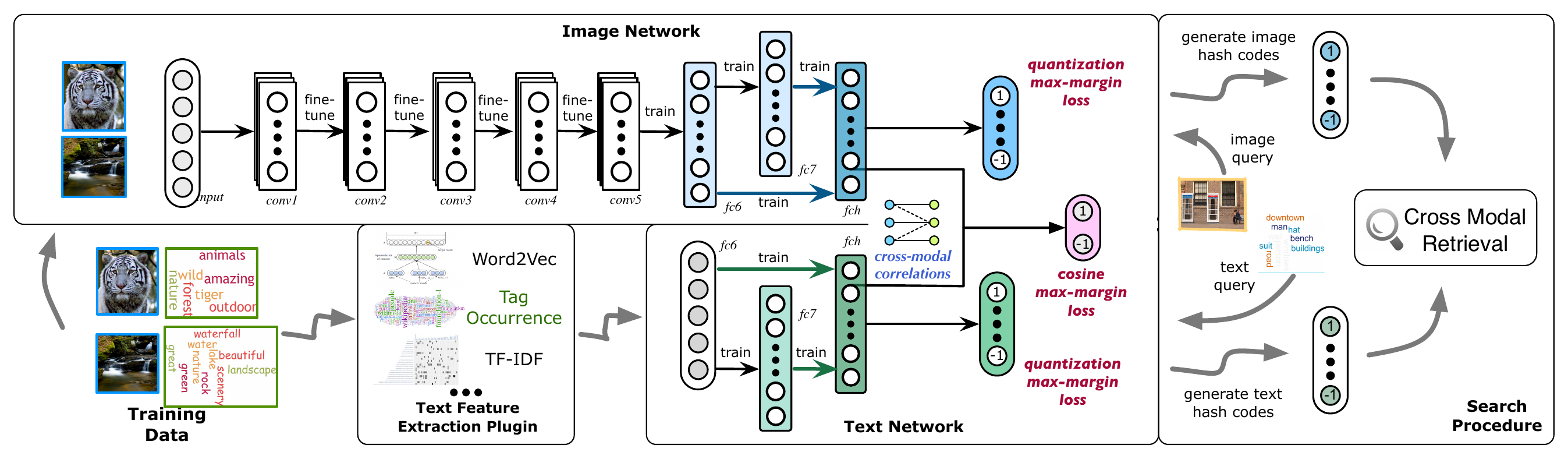}
  \vspace{-10pt}
  \caption{Correlation Hashing Network (CHN) for cross-modal retrieval, which constitutes (1) a convolutional neural network (CNN) for learning image representations, (2) a multilayer perceptions (MLP) for learning text representations, (3) two hashing layers $fch$ for generating hash codes, (4) a cosine max-margin loss for capturing cross-modal correlations, and a quantization max-margin loss for controlling hashing quality. The max-margin losses enhance the robustness to outlier points or hash bits.}
  \label{fig:CHN}
  \vspace{-10pt}
\end{figure*}

\section{Correlation Hashing Network}\label{section:CHN}
In cross-modal retrieval, the database consists of objects from one modality and the query consists of objects from another modality. We uncover the correlation structure underlying different modalities by learning from a training set of $n_x$ images $\{{\bm x}_i\}_{i=1}^{n_x}$ and $n_y$ texts $\{{\bm y}_j\}_{j=1}^{n_y}$, where ${\bm x}_i \in \mathbb{R}^{d_x}$ denotes the $d_x$-dimensional feature vector of the image modality, and ${\bm y}_j \in \mathbb{R}^{d_y}$ denotes the $d_y$-dimensional feature vectors of the text modality, respectively.
Some pairs of images and texts are associated with similarity labels $s_{ij}$, where $s_{ij} = 1$ implies ${\bm x}_i$ and ${\bm y}_j$ are similar and $s_{ij} = -1$ indicates ${\bm x}_i$ and ${\bm y}_j$ are dissimilar. In supervised hashing, $\mathcal{S} = \{s_{ij}\}$ can be constructed from the semantic labels of data points or the relevance feedback in click-through data. The goal of CHN is to jointly learn two modality-specific hashing functions $f_x\left( {\bm{x}} \right):{\mathbb{R}^{{d_x}}} \mapsto {\left\{ { - 1,1} \right\}^b}$ and $f_y\left( {\bm{y}} \right):{\mathbb{R}^{{d_y}}} \mapsto {\left\{ { - 1,1} \right\}^b}$ which respectively encode each unimodal point ${\bm x}$ and ${\bm y}$ in compact $b$-bit hash code ${\bm h}_x = f_x({\bm x})$ and ${\bm h}_y = f_y({\bm y})$ such that the similarity information conveyed in the given bimodal object pairs ${\cal S}$ is maximally preserved.

The Correlation Hashing Network (CHN) is a hybrid deep architecture for supervised learning to hash, as shown in Figure~\ref{fig:CHN}. The hybrid architecture accepts input in a pairwise form $({\bm x}_i, {\bm y}_j, s_{ij})$ and processes them through the deep representation learning and hash coding pipeline: (1) an image network with multiple convolution-pooling layers to extract good image representations, and a text network with several fully-connected layers to extract good text representations; (2) two fully-connected hashing layers to generate modality-specific compact hash codes; (3) a cosine max-margin loss for capturing cross-modal correlation; and (4) a quantization max-margin loss for controlling the quality of hash coding.

\subsection{Hybrid Deep Architecture}
The hybrid deep architecture for learning cross-modal hash functions are shown in Figure~\ref{fig:CHN}, which constitutes an image network and a text network. In the image network, we extend AlexNet \cite{cite:NIPS12CNN}, a deep convolutional neural network (CNN) comprised of five convolutional layers $conv1$--$conv5$ and three fully connected layers $fc6$--$fc8$. We replace the $fc8$ layer with a new $fch$ hash layer with $b$ hidden units, which transforms the network activation ${\bm u}_i$ in $b$-bit hash code by sign thresholding ${\bm h}^x_{i} = \operatorname{sgn} ({\bm u}_i)$. In text network, we adopt the Multilayer perceptrons (MLP) comprising three fully connected layers, of which the last layer is replaced with a new $fch$ hash layer with $b$ hidden units to transform the network activation ${\bm v}_i$ in $b$-bit hash code by sign thresholding ${\bm h}^y_{i} = \operatorname{sgn} ({\bm v}_i)$. We adopt the hyperbolic tangent (tanh) function to squash the activations to be within $[-1,1]$, which reduces the gap between the $fch$-layer representations ${\bm u}_i, {\bm v}_i$ and the binary hash codes ${\bm h}_i^{x}, {\bm h}_i^{y}$. We design new loss functions over the hash codes  generated by the deep networks for cross-modal correlation learning and quantization error minimization, which enable effective cross-modal retrieval.

\subsection{Cosine Max-Margin Loss}
\begin{figure}[tbp]
  \centering
  \includegraphics[width=0.9\columnwidth]{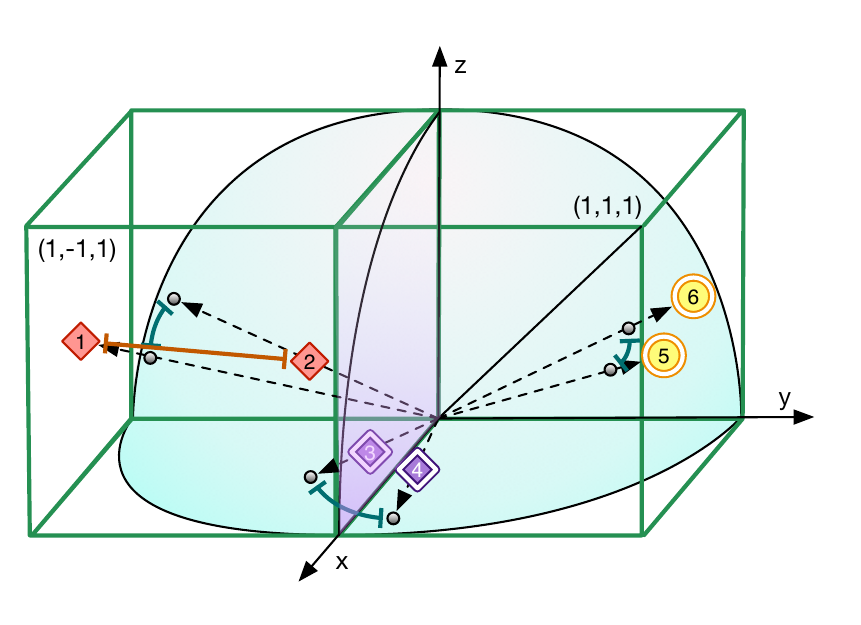}
  \vspace{-20pt}
  \caption{Motivation of the cosine max-margin loss and the quantization max-margin loss. (1) Similar points $1$ and $2$ (in red): large Euclidean distance (bad Hamming surrogate) but small cosine distance (better Hamming surrogate). (2) Similar points $3$ and $4$ (in purple): small cosine distance but large Hamming distance (the gap between cosine and Hamming). (3) Similar points $5$ and $6$ (in yellow): small cosine distance and small Hamming distance (the gap between cosine and Hamming is closed by the quantization max-margin loss).}
  \label{fig:CHN_Angular}
  \vspace{-10pt}
\end{figure}

For a pair of binary codes ${\bm h}^x_i$ and ${\bm h}^y_j$, there is a relationship between their Hamming distance $\mathrm{dist}_H(\cdot,\cdot)$ and their inner product $\langle \cdot,\cdot \rangle$: ${\textrm{dis}}{{\text{t}}_H}\left( {{{\bm{h}}^x_i},{{\bm{h}}^y_j}} \right) = \frac{1}{2}\left( {b - \left\langle {{{\bm{h}}^x_i},{{\bm{h}}^y_j}} \right\rangle } \right)$. Thus, we may use the inner product as a reasonable surrogate of the Hamming distance to quantify the pairwise similarity. However, note that ${\bm h}^x_i = \operatorname{sgn}({\bm u}_i)$ and ${\bm h}^y_i = \operatorname{sgn}({\bm v}_i)$, hence the approximation of such a surrogate for continuous representations ${\bm u}_i$ and ${\bm v}_j$ will be inaccurate if their vector lengths are very different, i.e. $\frac{1}{2}\left( {b - \left\langle {{{\bm{u}}_i},{{\bm{v}}_j}} \right\rangle } \right) \in (-\infty,+\infty)$ will no longer be a good surrogate of ${\textrm{dis}}{{\text{t}}_H}\left( {{{\bm{h}}^x_i},{{\bm{h}}^y_j}} \right) \in [-b,+b]$. Figure~\ref{fig:CHN_Angular} shows such a bad case, where points $1$ and $2$ (in red) have very different vector lengths and thus large Euclidean distance, but their Hamming distance is $0$ since they are assigned to the same binary code $(1,-1,1)$. The gap between Hamming distance and inner product has raised a serious misspecification issue of existing inner product based deep hashing methods \cite{cite:AAAI14CNNH,cite:CVPR15DNNH}.

To close the gap between Hamming distance and inner product for continuous representations, note that for a pair of binary codes ${\bm h}^x_i$ and ${\bm h}^y_j$, there is another relationship between their Hamming distance $\mathrm{dist}_H(\cdot,\cdot)$ and the cosine distance $\cos (\cdot,\cdot)$: ${\textrm{dis}}{{\text{t}}_H}\left( {{{\bm{h}}^x_i},{{\bm{h}}^y_j}} \right) = \frac{b}{2}\left( {1 - \cos\left( {{{\bm{h}}^x_i},{{\bm{h}}^y_j}} \right) } \right)$, where $\cos \left( {{{\bm{u}}_i},{{\bm{v}}_j}} \right) = \frac{\left\langle {{\bm{u}}_i,{\bm{v}}_j} \right\rangle} {\left\| {{\bm{u}}_i} \right\|\left\| {{\bm{v}}_j} \right\|}$, and $\|\cdot\|$ is the vector length. Since cosine distance can mitigate the diversity of vector lengths and make the continuous representations ${\bm u}_i$ and ${\bm v}_j$ lie on the unit sphere (which is important for cross-modal data as they usually have very different vector lengths), it makes $\frac{b}{2}\left( {1 - \cos \left( {{{\bm{u}}_i},{{\bm{v}}_j}} \right) } \right) \in [-b,+b]$ a more accurate surrogate of ${\textrm{dis}}{{\text{t}}_H}\left( {{{\bm{h}}^x_i},{{\bm{h}}^y_j}} \right)$ especially for comparing continuous representations of different modalities. As can be seen in Figure~\ref{fig:CHN_Angular}, the cosine distance between points $1$ and $2$ (in red) is close to zero and thus better approximates their Hamming distance. Hence in this paper, we opt to use the cosine distance as a good surrogate of the Hamming distance, which leads to new cosine-distance based structural loss functions.

To maximize the cross-modal correlation, we propose the following criterion: for each pair of objects $({\bm x}_i, {\bm y}_j, s_{ij})$, if $s_{ij} = 1$, indicating that ${\bm x}_i$ and $\bm{y}_j$ are similar, then their binary hash codes must be similar across different modalities, i.e. the Hamming distance should satisfy $d_H({\bm h}^x_i$, ${\bm h}^y_j) \rightarrow 0$, which implies the cosine distance should satisfy $\cos({\bm u}_i$, ${\bm v}_j) \rightarrow 1$. Correspondingly, if $s_{ij} = -1$, indicating that ${\bm x}_i$ and $\bm{y}_j$ are dissimilar, then by derivation, the cosine distance should satisfy $\cos({\bm u}_i$, ${\bm v}_j) \rightarrow -1$. It is very important to note that, for other widely-used distance metrics (e.g. inner product, Euclidean distance, etc), it is very difficult to devise such a well-specified learning criterion because these distances are not good surrogates of the Hamming distance. A straight-forward loss for achieving the above goal is the squared loss ${{\left( {{s_{ij}} - \cos \left( {{{\bm{u}}_i},{{\bm{v}}_j}} \right)} \right)}^2}$, however, the squared loss is not robust to outlier pairs of points. Motivated by SVMs, the similarity-preserving criterion leads to a novel cosine max-margin loss for maximizing cross-modal correlation as
\begin{equation}\label{eqn:CrossL}
  L = \sum\limits_{{s_{ij}} \in \mathcal{S}} {\max } \left( {0,{\delta } - {s_{ij}}\frac{{\left\langle {{{\bm{u}}_i},{{\bm{v}}_j}} \right\rangle }}{{\left\| {{{\bm{u}}_i}} \right\|\left\| {{{\bm{v}}_j}} \right\|}}} \right) ^2 ,
\end{equation}
where $0 < \delta \le 1$ is the margin parameter. The range of cosine distance ${\cos \left( {{{\bm{u}}_i},{{\bm{v}}_j}} \right)} \in [-1,1]$ is consistent with binary similarity labels $s_{ij}\in\{-1,1\}$, making the cosine max-margin loss in Equation~\eqref{eqn:CrossL} a well-specified loss for preserving the pairwise similarity information conveyed in $\mathcal{S}$. The cosine max-margin loss loss is powerful for cross-modal correlation analysis, since the vector lengths are very diverse in different modalities and may make other distance metrics (e.g. inner product) misspecified. In real retrieval systems, cosine distance is widely used to mitigate the diversity of vector lengths and significantly improve the retrieval quality, but to date, it has not been explored in deep hashing methods for cross-modal retrieval tasks \cite{cite:Arxiv14HashSurvey}. 

\subsection{Quantization Max-Margin Loss}
Though we justify that cosine distance is a good surrogate of Hamming distance, such an approximation may fail when two similar points ${\bm u}_i$ and ${\bm v}_j$ with $s_{ij} = 1$ (i.e. their cosine distance is small due to minimizing the cosine max-margin loss) lie on different sides of the hyperplane (i.e. their Hamming distance is large due to different signs of hash codes across the hyperplane). Figure~\ref{fig:CHN_Angular} shows such a failure case, where points $3$ and $4$ (in purple) have small cosine distance but large Hamming distance because they are assigned with different binary codes $(1,-1,1)$ and $(1,1,1)$, respectively. Such a gap between Hamming distance and cosine distance may result in an inaccurate surrogate approximation.

To close the gap between Hamming distance and cosine distance for continuous representations, note that for a pair of continuous representations ${\bm u}_i$ and ${\bm v}_j$, if they are close (in cosine distance) to their signed codes ${\bm h}^x_i = \operatorname{sgn}({\bm u}_i)$ and ${\bm h}^y_i = \operatorname{sgn}({\bm v}_i)$ (i.e. far from the hyperplane), then they will lie in the same hypercube with high probability (i.e. with the same binary code and hence their Hamming distance is zero). Instead of using the squared loss ${\left( {1 - \cos \left( {\left| {{{\bm{u}}_i}} \right|,{\bm{1}}} \right)} \right)^2}$ which is not robust to outlier bits especially for unbalanced encoding, we propose a new quantization max-margin loss
\begin{equation}\label{eqn:QuanL}
\textstyle{
{Q} = \sum\limits_{i = 1}^{n_x} {{\max \left( {0,{\delta } - \frac{{\left\langle {\left| {{{\bm{u}}_i}} \right|,{\bm{1}}} \right\rangle }}{{\left\| {{{\bm{u}}_i}} \right\|\left\| {\bm{1}} \right\|}}} \right)}}
+ \sum\limits_{i = 1}^{n_y} {{\max \left( {0,{\delta } - \frac{{\left\langle {\left| {{{\bm{v}}_i}} \right|,{\bm{1}}} \right\rangle }}{{\left\| {{{\bm{v}}_i}} \right\|\left\| {\bm{1}} \right\|}}} \right)}}}, \\
\end{equation}
where $0 < \delta \le 1$ is the margin parameter. Note that, minimizing the quantization max-margin loss will not only close the gap between the Hamming distance and cosine distance, but also lead to lower quantization error when binarizing the continuous representations ${\bm u}_i \in \mathbb{R}^b$ and ${\bm v}_j \in \mathbb{R}^b$ to hash codes ${\bm h}^x_i = \operatorname{sgn}({\bm u}_i)\in \{-1,1\}^b$ and ${\bm h}^y_j = \operatorname{sgn}({\bm v}_j) \in \{1,-1\}^b$, especially for unbalanced codes with outlier bits.

\subsection{Hash Function Learning}
We perform joint representation learning and hash coding by integrating Equations \eqref{eqn:CrossL}--\eqref{eqn:QuanL} in a joint optimization problem
\begin{equation}\label{eqn:Unified}
\mathop {\min }\limits_\Theta  O \triangleq L + \lambda Q,
\end{equation}
where $\Theta \triangleq \left\{ {{{\bm{W}}^\ell },{{\bm{b}}^\ell }} \right\}$ is the set of network parameters, $\lambda$ is the tradeoff parameter for the quantization max-margin loss. 
Finally, we can obtain $b$-bit binary codes by binarization using the sign function, ${\bm h}_x \leftarrow \mathrm{sgn}({\bm u})$ and ${\bm h}_y \leftarrow \mathrm{sgn}({\bm v})$, where $\forall i$, $\mathrm{sgn}(u_i) = 1$ if $u_i > 0$, otherwise $\mathrm{sgn}(u_i) = -1$. 

\begin{table*}[!htbp]
    \addtolength{\tabcolsep}{1.0pt} 
    \centering 
    \caption{Comparison of Mean Average Precision (MAP) on Two Cross-Modal Retrieval Tasks ($I \rightarrow T$ and $T \rightarrow I$)}
    \label{table:MAP}
    \small
    \begin{tabular}{c|c|cccc|cccc}
        \Xhline{1.0pt}
        \multirow{2}{20pt}{\centering Task} & \multirow{2}{20pt}{\centering Method} & \multicolumn{4}{c|}{NUS-WIDE} & \multicolumn{4}{c}{MIR-Flickr}\\
        \cline{3-10}
        & & 16 bits & 32 bits  & 64 bits  & 128 bits & 16 bits & 32 bits  & 64 bits  & 128 bits \\
        \hline
        \multirow{9}{30pt}{\centering $ I \rightarrow T$} 
& CVH \cite{cite:IJCAI11CVH} & 0.4454 & 0.4342 & 0.4290 & 0.4479 & 0.6883 & 0.7092 & 0.6976 & 0.6334\\
& IMH \cite{cite:SIGMOD13IMH} & 0.5256 & 0.6358 & 0.6151 & 0.6183 & 0.6765 & 0.6989 & 0.6964 & 0.6839\\
& CMSSH \cite{cite:CVPR10CMSSH} & 0.4665 & 0.4809 & 0.5670 & 0.5288 & 0.5122 & 0.5404 & 0.5842 & 0.5740\\
& RaHH \cite{cite:KDD13RAHH} & 0.6047 & 0.6312 & 0.6354 & 0.6534 & 0.6899 & 0.7086 & 0.7155 & 0.7204\\
& SCM \cite{cite:AAAI14SCM} & {0.6871} & {0.7271} & {0.7600} & {0.7739} & 0.6953 & 0.7091 & 0.7070 & 0.7497\\
& SePH \cite{cite:CVPR15SEPH} & 0.5982 & 0.5910 & 0.5988 & 0.6239 & {0.7526} & {0.7604} & {0.7607} & {0.7651}\\
& MMNN \cite{cite:TPAMI14MMNN} & 0.6255 & 0.6424 & 0.6514 & 0.6713 & 0.6915 & 0.7185 & 0.7277 & 0.7352\\
& DCMH \cite{cite:ARXIV16DCMH} & \underline{0.7353} & \underline{0.7628} & \underline{0.7805} & \underline{0.7912} & \underline{0.7576} & \underline{0.7985} & \underline{0.8152} & \underline{0.8369}\\
        \cline{2-10} 
& CHN & \textbf{0.7995} & \textbf{0.8146} & \textbf{0.8353} & \textbf{0.8662} & \textbf{0.8223} & \textbf{0.8477} & \textbf{0.8777} & \textbf{0.8808} \\
        \hline
        \multirow{9}{30pt}{\centering $ T \rightarrow I$} 
& CVH \cite{cite:IJCAI11CVH} & 0.4357 & 0.4253 & 0.4186 & 0.4184 & 0.6065 & 0.6277 & 0.6063 & 0.6004\\
& IMH \cite{cite:SIGMOD13IMH} & 0.6253 & 0.6816 & 0.7094 & 0.6532 & 0.6229 & 0.6201 & 0.6239 & 0.6237\\
& CMSSH \cite{cite:CVPR10CMSSH} & 0.4166 & 0.5110 & 0.4343 & 0.4974 & 0.4656 & 0.4624 & 0.4769 & 0.5337\\
& RaHH \cite{cite:KDD13RAHH} & 0.5786 & 0.6158 & 0.6214 & 0.6240 & 0.6248 & 0.6321 & 0.6359 & 0.6464\\
& SCM \cite{cite:AAAI14SCM} & {0.6794} & \underline{0.7194} & \underline{0.7480} & {0.7466} & 0.6173 & 0.6115 & 0.6177 & 0.6564\\
& SePH \cite{cite:CVPR15SEPH} & 0.6044 & 0.6036 & 0.6256 & 0.6405 & 0.6470 & 0.6429 & 0.6517 & 0.6550\\
& MMNN \cite{cite:TPAMI14MMNN} & 0.6083 & 0.6226 & 0.6435 & 0.6648 & {0.6815} & {0.6992} & {0.7082} & {0.7171}\\
& DCMH \cite{cite:ARXIV16DCMH} & \underline{0.6898} & 0.7102 & 0.7358 & \underline{0.7557} & \underline{0.7013} & \underline{0.7288} & \underline{0.7458} & \underline{0.7698}\\
        \cline{2-10}
& CHN & \textbf{0.7533} & \textbf{0.7803} & \textbf{0.7888} & \textbf{0.8288} & \textbf{0.7749} & \textbf{0.7891} & \textbf{0.8169} & \textbf{0.8258} \\
        \Xhline{1.0pt}
    \end{tabular}
    \normalsize
\end{table*}

\subsection{Learning Algorithm}
We derive learning algorithms for CHN in Equation~\eqref{eqn:Unified}, and show rigorously that both cosine max-margin loss and quantization max-margin loss can be optimized efficiently via standard back-propagation (BP) algorithm. For brevity, we define the pointwise cost of the image modality (the pointwise cost of the text modality is the same and omitted) as ${O_i^x} \triangleq \sum\nolimits_{j:{s_{ij}} \in {\mathcal S}} {{L_{ij}}}  + \lambda  Q_i^x$. We derive the gradient of point-wise cost $O^x_i$ w.r.t. ${\bm W}_{x,k}^{\ell}$, the network parameter of the $k$-th unit in the $\ell$-th layer for the image network as
\begin{equation}
 \small
 \begin{aligned}
  \frac{{\partial {O_i^x}}}{{\partial {\bm{W}}_{x,k}^\ell }}&  = \sum\limits_{j:{s_{ij}} \in {\mathcal S}} {\frac{{\partial {L_{ij}}}}{{\partial {\bm{W}}_{x,k}^\ell }}}  + \lambda \frac{{\partial Q_i^x}}{{\partial {\bm{W}}_{x,k}^\ell }}  \\
  &  = \left( {\sum\limits_{j:{s_{ij}} \in {\mathcal S}} {\frac{{\partial {L_{ij}}}}{{\partial \hat u_{ik}^\ell }}}  + \lambda \frac{{\partial Q_i^x}}{{\partial \hat u_{ik}^\ell }}} \right)\frac{{\partial \hat u_{ik}^\ell }}{{\partial {\bm{W}}_{x,k}^\ell }}  \\ 
  &  = \delta _{x,ik}^\ell {\bm{u}}_i^{\ell  - 1} , \\
 \end{aligned}
 \normalsize
\end{equation}
where ${\bm{\hat u}}_{i}^{\ell}  = {{{\bm{W}}_x^{\ell}}{\bm{u}}_{i}^{\ell  - 1} + {{\bm{b}}_x^{\ell}}}$ is $\ell$-th layer output before activation $a_x^{\ell}(\cdot)$, $\delta _{x,ik}^{\ell}  \triangleq \sum\nolimits_{j:{s_{ij}} \in {\mathcal S}} {\frac{{\partial {L_{ij}}}}{{\partial \hat u_{ik}^\ell }}}  + \lambda \frac{{\partial Q_i^x}}{{\partial \hat u_{ik}^\ell }}$
 is the point-wise \emph{residual} term that measures how much the $k$-th unit in the $\ell$-th layer is responsible for the error of point ${\bm x}_i$ in the network output. For an output unit $k$, we can measure the difference between the network's activation and the true target value, and use that to define the residual $\delta _{x,ik}^{l}$ as
\begin{equation}\label{eqn:deltaL}
  \scriptsize
  \begin{aligned}
  & \delta _{x,ik}^l = \sum\limits_{j: {s_{ij}} \in {\mathcal S}}  2 \cdot {\max } \left( {0,{\delta } - {s_{ij}}\frac{{\left\langle {{{\bm{u}}_i},{{\bm{v}}_j}} \right\rangle }}{{\left\| {{{\bm{u}}_i}} \right\|\left\| {{{\bm{v}}_j}} \right\|}}} \right) \\
  & \cdot \mathbb{I}\left( {{\delta} - {s_{ij}}\frac{{{\bm{u}}_i^l \cdot {\bm{v}}_j^l}}{{\left\| {{\bm{u}}_i^l} \right\|\left\| {{\bm{v}}_j^l} \right\|}} > 0} \right) \dot a_x^l\left( {\hat u_{ik}^l} \right) \\
  & \cdot \left[ { - {s_{ij}}\left( {\frac{{v_{jk}^l}}{{\left\| {{\bm{u}}_i^l} \right\|\left\| {{\bm{v}}_j^l} \right\|}} - \frac{{u_{ik}^l\left\langle {{\bm{u}}_i^l,{\bm{v}}_j^l} \right\rangle }}{{{{\left\| {{\bm{u}}_i^l} \right\|}^3}\left\| {{\bm{v}}_j^l} \right\|}}} \right)} \right] - \lambda \dot a_x^l\left( {\hat u_{ik}^l} \right) \\
  & \cdot \mathbb{I}\left( {{\delta} - \frac{{\sum\nolimits_{j = 1}^b {\left| {{u^l_{ij}}} \right|} }}{{\sqrt b \left\| {{{\bm{u}}^l_i}} \right\|}} > 0} \right)  \; \left[ {\frac{{\operatorname{sgn} \left( {u_{ik}^l} \right)}}{{\sqrt b \left\| {{\bm{u}}_i^l} \right\|}} - \frac{{u_{ik}^l {\sum\nolimits_{j = 1}^b {\left| {{u^l_{ij}}} \right|} } }}{{\sqrt b {{\left\| {{\bm{u}}_i^l} \right\|}^3}}}} \right] \\
	\end{aligned}
	\normalsize
\end{equation}
where ${\dot a}_x^{l}(\cdot)$ is the derivative of the $l$-th layer activation function, and $\mathbb{I}(A)$ is an indicator function, $\mathbb{I}(A) = 1$ if $A$ is true and $\mathbb{I}(A) = 0$ otherwise. For a hidden unit $k$ in the $(\ell-1)$-th layer, we compute residual $\delta _{x,ik}^{\ell-1}$ based on a weighted average of the errors of all units ${k'}=1,\ldots,n_{\ell-1}$ in the $(\ell-1)$-th layer that use ${\bm u}_i^{\ell-1}$ as an input, which is consistent with BP,
\begin{equation}\label{eqn:deltaEll}
	\small
    \delta _{x,ik}^{\ell  - 1} = \left( {\sum\limits_{k' = 1}^{{n_{\ell  - 1}}} {\delta _{x,ik'}^{\ell} W_{x,kk'}^{\ell  - 1}} } \right){{\dot a}_x^{\ell  - 1}}\left( {\hat u_{ik}^{\ell  - 1}} \right) ,
\end{equation}
where $n_{\ell-1}$ is number of units in $(\ell-1)$-th layer. The residuals in all other layers can be computed by back-propagation. The overall computational complexity is $O(|\mathcal{S}|)$, where $|\mathcal{S}|$ is the number of cross-modal similarity pairs in $\mathcal{S}$ for training.


\section{Experiments}\label{section:Experiments}

\subsection{Setup}
\textbf{NUS-WIDE} \cite{cite:CIVR09NusWide} is a web image dataset of 81 ground truth concepts manually annotated for evaluation. Following prior works \cite{cite:VLDB14MSAE,cite:MM13LCMH}, we use the subset of 195,834 image-text pairs that belong to some of the 21 most frequent concepts. All images are resized into 256$\times$256.
\textbf{MIR-Flickr} \cite{cite:ICMR08Flickr1M} consists of 25,000 images collected from the Flickr website, where each image is labeled with some of 38 semantic concepts. We resize images of this labeled subset into 256$\times$256. 

For our deep learning based approach CHN, we directly use the raw image pixels as the input. 
For fair comparison, for traditional {shallow} hashing methods, we use AlexNet \cite{cite:NIPS12CNN} to extract deep \emph{fc7} features for each image in two benchmark datasets by a 4096-dimensional vector.
For text modality, all the methods use tag occurrence vectors as the input. In NUS-WIDE, we randomly select 100 pairs per class as the query set, 500 pairs per class as the training set and 50 pairs per class as the validation set. In MIR-Flickr, we randomly select 1000 pairs as the query set, 4000 pairs as the training set and 1000 pairs as the validation set. The similarity pairs for training are constructed using semantic labels: each pair is similar (dissimilar) if they share at least one (none) semantic label.

\begin{figure*}[tbp]
    \centering
    \subfigure[\scriptsize{$ I \rightarrow T$ on NUS-WIDE}]{
        \includegraphics[width=0.22\textwidth]{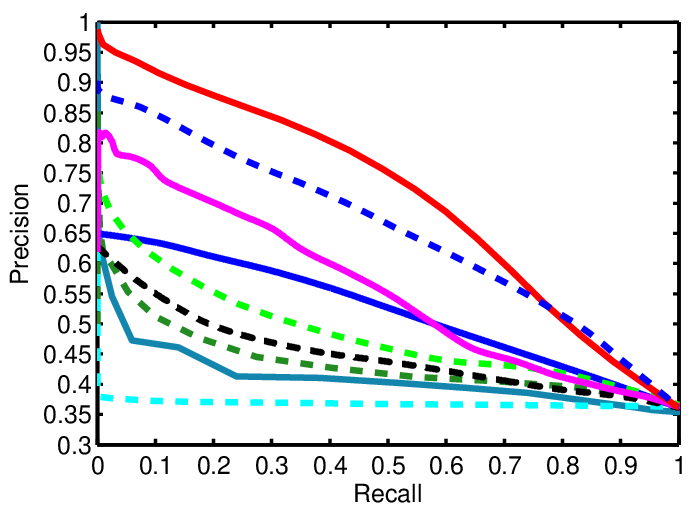}
        \label{fig:pr_nuswide_2}
    }
    \subfigure[\scriptsize{$ T \rightarrow I$ on NUS-WIDE}]{
        \includegraphics[width=0.22\textwidth]{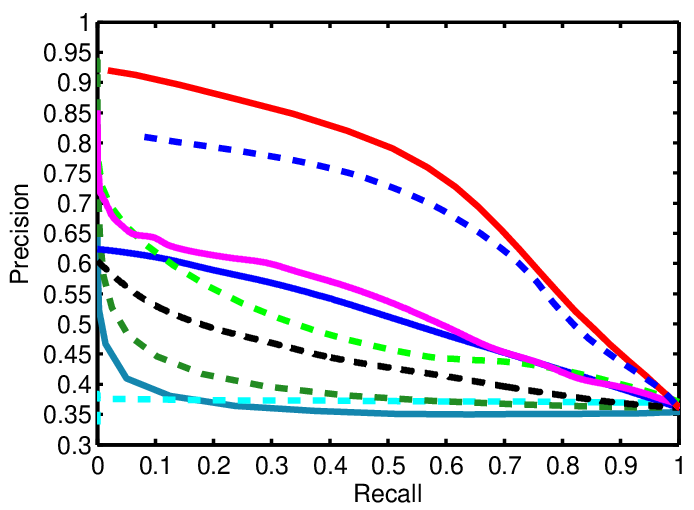}
        \label{fig:pr_nuswide_4}
    }
    \subfigure[\scriptsize{$ I \rightarrow T$ on MIR-Flickr}]{
        \includegraphics[width=0.22\textwidth]{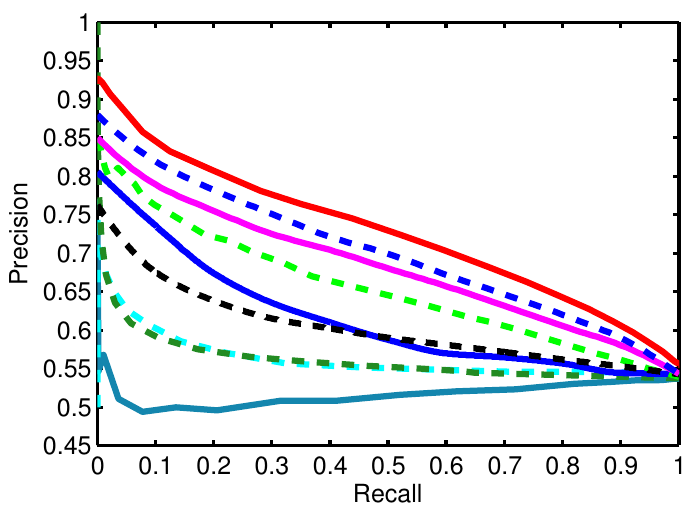}
        \label{fig:pr_flickr1m_2}
    }
    \subfigure[\scriptsize{$ T \rightarrow I$ on MIR-Flickr}]{
        \includegraphics[width=0.282\textwidth]{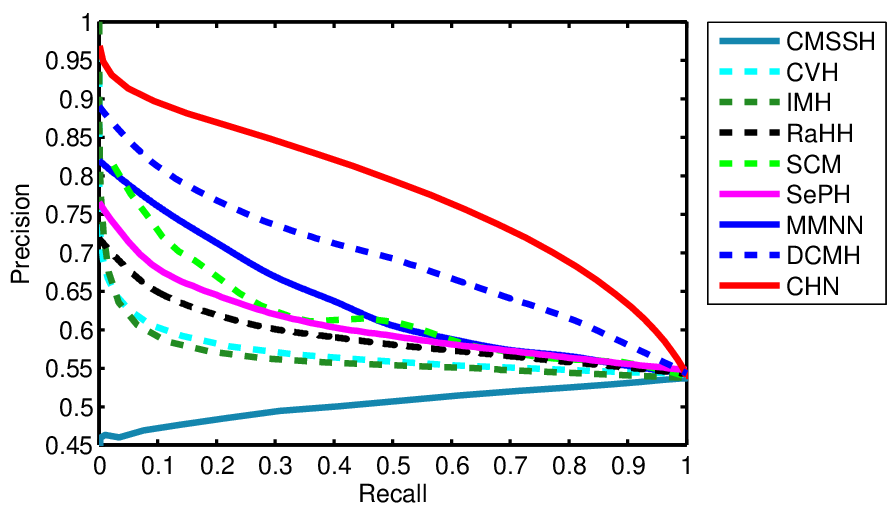}
        \label{fig:pr_flickr1m_4}
    }\\\vspace{-8pt}
    \subfigure[\scriptsize{$ I \rightarrow T$ on NUS-WIDE}]{
        \includegraphics[width=0.22\textwidth]{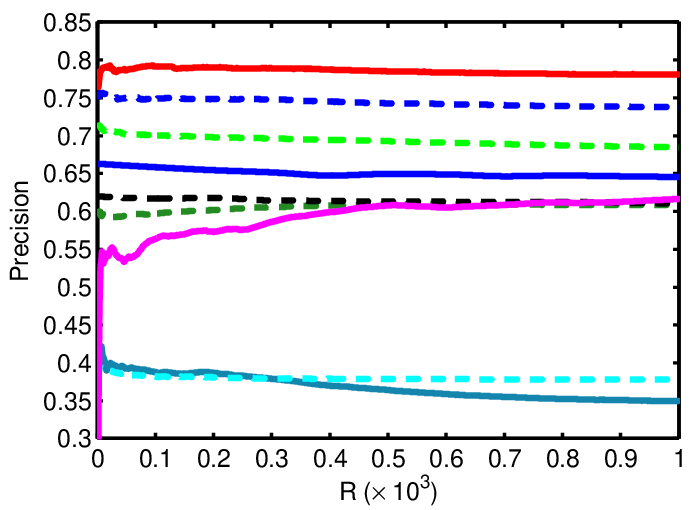}
        \label{fig:prec_nuswide_2}
    }
    \subfigure[\scriptsize{$ T \rightarrow I$ on NUS-WIDE}]{
        \includegraphics[width=0.22\textwidth]{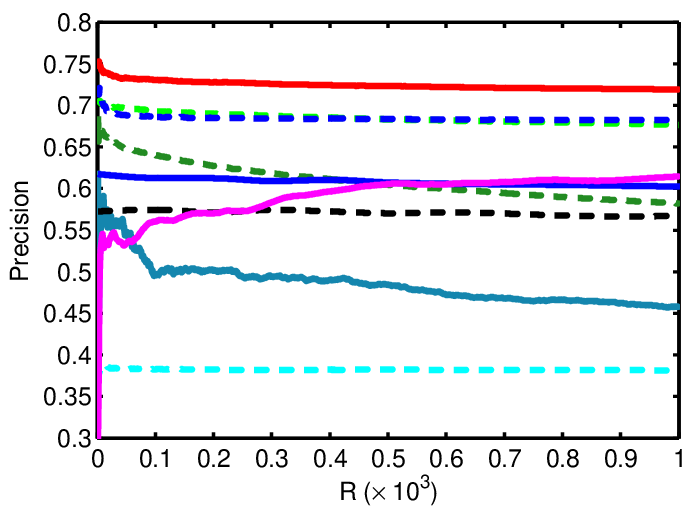}
        \label{fig:prec_nuswide_4}
    }
    \subfigure[\scriptsize{$ I \rightarrow T$ on MIR-Flickr}]{
        \includegraphics[width=0.22\textwidth]{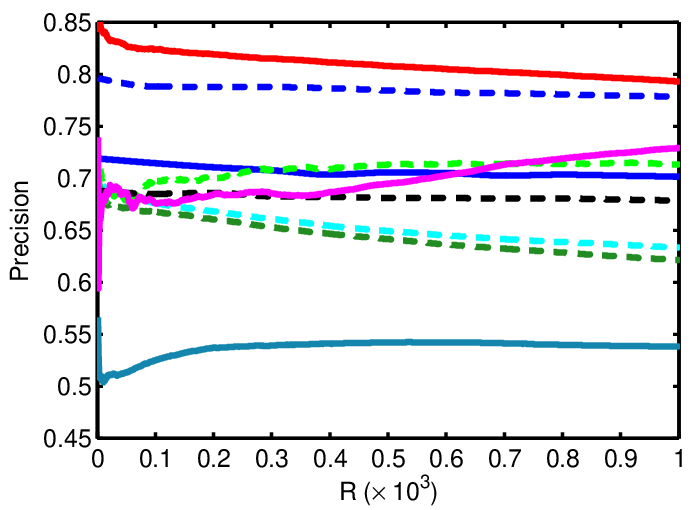}
        \label{fig:prec_flickr1m_2}
    }
    \subfigure[\scriptsize{$ T \rightarrow I$ on MIR-Flickr}]{
        \includegraphics[width=0.282\textwidth]{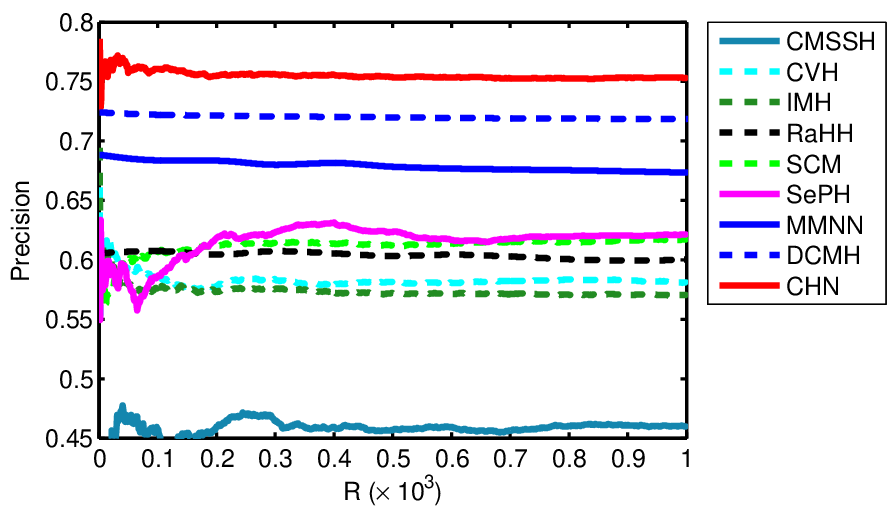}
        \label{fig:prec_flickr1m_4}
    }
	\vspace{-10pt}
    \caption{Precision-recall curves (a)-(d) and Precision@top R curves (e)-(h) on NUS-WIDE and MIR-Flickr with 32 bits codes.}
    \label{fig:pr}
    \vspace{-10pt}
\end{figure*}

We compare CHN with state-of-the-art cross-modal hashing and deep hashing methods, including three unsupervised methods 
\textbf{IMH} \cite{cite:SIGMOD13IMH}, 
\textbf{CVH} \cite{cite:IJCAI11CVH}, and 
\textbf{MMNN} \cite{cite:TPAMI14MMNN}, 
and five supervised methods 
\textbf{CMSSH} \cite{cite:CVPR10CMSSH}, 
\textbf{RaHH} \cite{cite:KDD13RAHH}, 
\textbf{SCM} \cite{cite:AAAI14SCM}, 
\textbf{SePH} \cite{cite:CVPR15SEPH} and 
\textbf{DCMH} \cite{cite:ARXIV16DCMH}, 
where \textbf{MMNN} and \textbf{DCMH} are deep learning based methods. 
We follow \cite{cite:AAAI14SCM,cite:ARXIV16DCMH} to evaluate retrieval quality via three standard metrics: Mean Average Precision (MAP), precision-recall curves and precision@top-R curves. 

We implement the CHN model based on the open-source \textbf{Caffe} framework \cite{cite:MM14Caffe}. 
For image network, we employ the AlexNet architecture \cite{cite:NIPS12CNN}, fine-tune $conv1$--$fc7$ that were copied from the pre-trained model, and train hashing layer $fch$, all via back-propagation.
For text network, we employ a two layer multi-layer perceptrons (MLP), in which the $fc7$ layer has 4096 ReLU units with dropout rate 0.5, and the $fch$ layer have $b$ $\rm{tanh}$ units.
We use mini-batch SGD with 0.9 momentum, and cross-validate the learning rate from $10^{-5}$ to $1$ with a multiplicative step-size $10$, and fix mini-batch size as $64$. For all methods, we select their parameters using cross-validation. Each experiment repeats five runs and average results are reported.

\begin{table*}[htb]
    \addtolength{\tabcolsep}{5pt} 
    \centering 
    \caption{Mean Average Precision (MAP) of CHN Variants on NUS-WIDE and MIR-Flickr Datasets}
    \label{table:EmpirMAP}
    \small
    \begin{tabular}{c|c|cccc|cccc}
        \Xhline{1.0pt}
        \multirow{2}{30pt}{\centering Task} & \multirow{2}{30pt}{\centering Method} & \multicolumn{4}{c|}{NUS-WIDE} & \multicolumn{4}{c}{MIR-Flickr}\\
        \cline{3-10}
        & & 16 bits & 32 bits  & 64 bits  & 128 bits & 16 bits & 32 bits  & 64 bits  & 128 bits \\
        \hline
        \multirow{6}{30pt}{\centering $ I \rightarrow T$} 
 & CHN-M & 0.6987 & 0.7251 & 0.7350 & 0.7593 & 0.7313 & 0.7665 & 0.8148 & 0.8281 \\
 & CHN-I & 0.6751 & 0.7047 & 0.7214 & 0.7253 & 0.7105 & 0.7481 & 0.7798 & 0.7843 \\
 & CHN-Q & 0.7743 & 0.7982 & 0.8212 & 0.8598 & 0.7893 & 0.8214 & 0.8558 & 0.8678 \\
 & CHN & \underline{0.7995} & \underline{0.8146} & \underline{0.8353} & \underline{0.8662} & \underline{0.8223} & \underline{0.8477} & \underline{0.8777} & \underline{0.8808} \\
 & CHN-B & \textbf{0.8650} & \textbf{0.8706} & \textbf{0.8746} & \textbf{0.8879} & \textbf{0.8753} & \textbf{0.8612} & \textbf{0.8905} & \textbf{0.8933} \\										
        \hline
        \multirow{6}{30pt}{\centering $ T \rightarrow I$} 
 & CHN-M & 0.5789 & 0.5924 & 0.5997 & 0.6318 & 0.6532 & 0.6875 & 0.7015 & 0.7135 \\
 & CHN-I & 0.5874 & 0.6012 & 0.6241 & 0.6534 & 0.6817 & 0.6987 & 0.7314 & 0.7389 \\
 & CHN-Q & 0.7395 & 0.7543 & 0.7779 & 0.8053 & 0.7515 & 0.7687 & 0.8043 & 0.8125 \\
 & CHN & \underline{0.7533} & \underline{0.7803} & \underline{0.7888} & \underline{0.8288} & \underline{0.7749} & \underline{0.7891} & \underline{0.8169} & \underline{0.8258} \\
 & CHN-B & \textbf{0.8003} & \textbf{0.8095} & \textbf{0.8185} & \textbf{0.8407} & \textbf{0.7915} & \textbf{0.8142} & \textbf{0.8207} & \textbf{0.8308} \\
        \Xhline{1.0pt}
    \end{tabular}
    \normalsize
\end{table*}

\subsection{Results}
We report in Table \ref{table:MAP} the MAP of all methods with different code lengths, i.e. 16, 32, 64 and 128 bits. CHN substantially outperforms all state-of-the-art methods for all cross-modal retrieval tasks. 
Specifically, for NUS-WIDE dataset, CHN outperforms the best shallow method SCM by 9.19\% / 6.44\% in average MAP for $I \rightarrow T$ / $T \rightarrow I$.
For MIR-Flickr dataset, CHN outperforms the best shallow method SePH by 9.74\% / 15.25\% in average MAP for $I \rightarrow T$ / $T \rightarrow I$. 
Compared to deep cross-modal hashing methods, CHN outperforms state-of-the-art DCMH by large margins of 6.15\% / 6.49\% and 5.51\% / 6.53\%. 
These results verify that CHN is able to learn high-quality hash codes for effective cross-modal retrieval.

We respectively report in Figure \ref{fig:pr} (a)-(d) the precision-recall curves with 32 bits for two cross-modal retrieval tasks $ I \rightarrow T$ and $ T \rightarrow I$ on two benchmark datasets NUS-WIDE and MIR-Flickr. CHN shows the best retrieval performance at all recall levels.
Figure \ref{fig:pr} (e)-(h) respectively show the precision@top-R curves of all state-of-the-art methods, which further represent the precision changes along with the number of top-R retrieved results ($R = 1000$) with 32 bits on NUS-WIDE and MIR-Flickr datasets. CHN significantly outperforms all state-of-the-art methods under these metrics. 



\subsection{Discussion}\label{section:Empirical}

To go deeper with the efficacy of CHN, we investigate four variants of CHN: 
\textbf{CHN-M} is the CHN variant without using the margin parameter, in other words, $\delta = 1.0$;
\textbf{CHN-I} is the CHN variant that replaces the cosine max-margin loss \eqref{eqn:CrossL} with the widely-used inner-product squared loss $L = \sum\nolimits_{{s_{ij}} \in \mathcal{S}} {{{\left( {{s_{ij}} - \frac{1}{K}\left\langle {{\bm{h}}_i,{\bm{h}}_j} \right\rangle } \right)}^2}}$\cite{cite:CVPR12KSH,cite:AAAI14CNNH}; 
\textbf{CHN-Q} is the CHN variant without using the quantization max-margin loss \eqref{eqn:QuanL};
\textbf{CHN-B} is the CHN variant without using binarization on hash codes, which may serve as the upper bound of retrieval performance. 

From Table~\ref{table:EmpirMAP}, we have the following key observations.
\textbf{(a)} CHN outperforms CHN-M by large margins, demonstrating that the max-margin principle can significantly enhance the robustness of the hash codes to the outlier points. 
\textbf{(b)} By using the cosine max-margin loss, CHN outperforms CHN-I by large margins. The squared inner-product loss has been widely adopted in the previous works \cite{cite:AAAI14CNNH,cite:CVPR12KSH}. However, this loss cannot link well the pairwise distances between continuous representations (taking values in $(-\infty,+\infty)$ when using continuous relaxation) to the pairwise similarity labels (taking binary values \{-1,1\}). In contrast, the proposed cosine max-margin loss \eqref{eqn:CrossL} is inherently consistent with the training pairs. Besides, the margin parameter $\delta$ can also control the robustness level of the similarity-preserving procedure to the outlier points. 
The promising performance of CHN suggests that the proposed cosine max-margin loss can preserve cross-modal correlations and is well-specified to cross-modal retrieval scenarios. 
\textbf{(c)} By using quantization max-margin loss \eqref{eqn:QuanL}, CHN incurs small MAP decreases than CHN-Q when quantizing continuous representations into binary codes. Especially for shorter length of hash codes (16 bits), CHN-Q incurs huge decreases while CHN incurs negligible MAP decreases. This validates that quantization max-margin loss can effectively reduce the quantization error and obtain high-quality hash codes.
The experimental results also imply that all components in CHN are important for achieving the promising performance, and missing any component will lead to huge performance drop.


%

\section{Conclusion}\label{section:Conclusion}
In this paper, we have proposed a novel Correlation Hashing Network (CHN) for effective and efficient cross-modal retrieval. CHN is a hybrid deep architecture that jointly optimizes the new cosine max-margin loss on semantic similarity pairs and the new quantization max-margin loss on compact hash codes. Extensive experiments on standard cross-modal retrieval datasets show that the CHN model yields substantial boosts over the state-of-the-art hashing methods.

\footnotesize
\bibliographystyle{named}
\bibliography{HHN}

\end{document}